\documentclass{article}
\usepackage[nonatbib,final]{neurips_2020_ml4ad}
\usepackage[utf8]{inputenc} % allow utf-8 input
\usepackage[T1]{fontenc}    % use 8-bit T1 fonts
\usepackage{hyperref}       % hyperlinks
\usepackage{url}            % simple URL typesetting
\usepackage{booktabs}       % professional-quality tables
\usepackage{amsfonts}       % blackboard math symbols
\usepackage{nicefrac}       % compact symbols for 1/2, etc.
\usepackage{microtype}      % microtypography

\usepackage[numbers]{natbib}
\usepackage{amsmath,amssymb}       % blackboard math symbols
\usepackage[pdftex]{graphicx}
\usepackage[table,x11names]{xcolor} % added table for \rowcolor

\newcommand{\acro}{BEEF}
\newcommand{\hdd}{HDD}
\newcommand{\bddx}{BDD-X}
\newcommand{\eg}{e.g.\ }
\newcommand{\ie}{i.e.\ }

\newcommand{\tens}{\mathcal}
\newcommand{\mat}{\mathbf}
\newcommand{\bleu}{BLEU-4 }
\newcommand{\meteor}{METEOR }
\newcommand{\cider}{CIDEr-D }
\DeclareMathOperator{\dcnn}{3DCNN}
\DeclareMathOperator{\gru}{GRU}

\usepackage{makecell} % to have \makecell
\usepackage{tabularx} % to have tabularx
\usepackage{multirow} % to have \multirow
\usepackage{floatrow} % to put a figure next to a table
\usepackage{subcaption} % To have subcaptions for multiple subfigures
\title{Driving Behavior Explanation with Multi-level Fusion}
\author{H\'edi Ben-Younes\thanks{equal contribution. Correspondance to \texttt{hedi.ben-younes@valeo.com} and \texttt{eloi.zablocki@valeo.com}} \\
        Valeo.ai \\
        \And
        \'Eloi Zablocki\footnotemark[1] \\
        Valeo.ai \\
        \And
        Patrick P\'erez \\
        Valeo.ai \\
        \And 
        Matthieu Cord \\
        Valeo.ai \\
        Sorbonne Universit\'e \\
        }

\begin{document}

\maketitle

\begin{abstract}
In this era of active development of autonomous vehicles, it becomes crucial to provide driving systems with the capacity to explain their decisions.
In this work, we focus on generating high-level driving explanations as the vehicle drives.
We present \acro{}, for BEhavior Explanation with Fusion, a deep architecture which explains the behavior of a trajectory prediction model.
Supervised by annotations of human driving decisions justifications, \acro{} learns to fuse features from multiple levels.
Leveraging recent advances in the multi-modal fusion literature, \acro{} is carefully designed to model the correlations between high-level decisions features and mid-level perceptual features.
The flexibility and efficiency of our approach are validated with extensive experiments on the HDD and BDD-X datasets. 
\end{abstract}

\section{Introduction}
\vspace{-0.1cm}

% CV trivia
% It has shown \hedih{incredible} performance in 2D and 3D object detection \hedih{[refs]}, semantic segmentation \hedih{[refs]}, depth estimation \hedih{[refs]}, domain adaptation \hedih{[refs]}. 
% % Driving paradigms
% Even more challenging is the capacity of networks to learn directly how to drive, by imitating driving demonstrations \hedih{[refs]} or from trial-and-error through simulation \hedih{[refs]}.
% \eloi{Dernière phrase à mettre au début de paragraphe suivant ?}

% INTELLIGENT VEHICLES AND EXPLANABILITY
% Intelligent cars exist
Over the last decade, research communities have devoted lots of effort on making cars safer, reliable and intelligent.
% They are doing well
Vehicles are more and more able to understand their environment, plan motions and take appropriate decisions.
% Good accuracy: not enough
However, reaching a perfect driving accuracy is not sufficient to allow for real-world deployment of self-driving vehicles. % as it is a highly critical application.
% Need for explanability
In such a safety-critical domain, providing human-understandable explanations is crucial as it can favor the adoption of self-driving systems \cite{bddx,edmonds2019tale} by increasing users' trust \citep{trusthci20} and easing regulation and model validation \cite{beaudouin2020identifying,deeptest}.
% as examining explanations could facilitate the task of validating models and correcting failure modes \cite{deeptest}.
%Indeed, automated driving being a highly critical application, it is crucial that these systems can \emph{explain} their decisions in a human-understandable form. 
%However, in applications such as autonomous driving, providing human-understandable explanations is crucial as it can favor adoption of self-driving systems \cite{bddx,edmonds2019tale}.
%Moreover, examining the explanations for an autonomous vehicle could ease the task of validating models and correcting failure modes.
 
% DETECTION OF DRIVING CAUSES
% Besides learning directly how to drive, automatically analyzing driving situations has emerged as a key topic to design intelligent mobility applications. 
% Recent research attempts to learn models that recognize driver activity \cite{DBLP:journals/tvt/XingLWCVW19} and behavior \cite{dbus,DBLP:journals/corr/abs-1911-02172}, detect driver drowsiness \cite{drowsiness,DBLP:conf/mobisys/YouLCWCBMCLTC13} or automatically trigger emergency-braking \cite{DBLP:conf/itsc/CoelinghEB10}. 

% THE AIM OF THE PAPER, WHAT IS AN EXPLANATION ?
% The task
In this paper, we aim at explaining the decisions of a driving model to a human user, as the vehicle drives.
% Why it is useful: towards a long-term goal
This is grounded towards the long-term goal of building driving systems capable of communicating the underlying reasons behind their decisions. 
%\matt{j ai mis en comments la suite}
%This is challenging for two main reasons.
%% 1: online and not offline
%First, explanations should be given while the car drives, \ie in an \emph{online} fashion. This is in opposition to offline scenarios where either future information can be accessed \cite{li2020learning} or where sequences of interest are already segmented \cite{bddx}.
%% 2: self explanation and not rationalization
%Second, 
% What is an explanation? def is not obvious
Specifying the requirements an ``explanation'' should meet is a non-trivial problem per se \cite{DBLP:conf/dsaa/GilpinBYBSK18}.
% Saliency maps ?
Visual saliency maps \cite{saliencymaps,gradcam,gradcam++} constitute broadly accepted explanations for computer vision systems.
% What they are
They explain the network's predictions by highlighting spatial locations in the image on which the network relied the most to take its decision \cite{DBLP:journals/corr/BojarskiYCCFJM17,causal_attention}.
% Limits of saliency
However, saliency maps usually need to be interpreted, and their purely visual nature may not be well suited for human-machine interactions. 
% Then other type: language
As an answer, natural language justifications have been used to explain self-driving decisions \cite{bddx}.
% We also take inspiration from human driver cause recognition
Related to this effort on explainability, recent works leverage visual recognition methods to find justifications of \emph{human} drivers' decisions \cite{hdd_cvpr2018,li2020learning}. 
% Besides, in the context of driving, several works tackle the prediction of causes of \emph{human} driving, framed as a classifiction task \cite{hdd_cvpr2018,li2020learning}.
%\eloi{eventuellement parler des autres taches de recognition en driving (driver drowsiness...) cf ancien RW}
% We want introspective explanations
Most valuable explanations are \textit{introspective} in that they depend on the actual driving system and on its inner processing that leads to the produced driving decisions \cite{bddx}; this contrasts with post-hoc rationalizations that only consider the final driving decisions of the system \cite{DBLP:conf/iccv/LuoWQ13,DBLP:conf/cvpr/VemulapalliAC14,li2020learning}.

On top of a simple self-driving backbone, we design an \emph{explanation module} that produces introspective explanations by conditioning its reasoning on both the driving decisions and intermediate perceptual features of the driving system.
% Motivation: problem various explanations can collapse to the same output
This is motivated by the fact that final driving predictions hardly contain the necessary information to recover the precise cause for a decision, as various possible explanations collapse to the same behavior.
% We give an example for the reader to breathe
For example, a car braking can be caused by multiple factors such as a red light, crossing pedestrians or a stop sign. The precise cause is lost through the processing as these different situations trigger the same braking behavior.
% Insist on the need for intermediate features
We thus argue that features at different semantic levels need to be fused together: late features containing the system decisions, in the form of the predicted trajectory, and intermediate spatio-temporal features containing perceptual scene information.
% Thus emerging problem for the fusion
A \emph{fusion} problem thus emerges to aggregate information from these multi-level features.
% How we solve it
In that respect, we leverage the recent literature on multi-modal fusion techniques for Visual Question Answering (VQA) systems \cite{benyounescadene2017mutan,ben2019block} that we adapt to our needs.

\begin{figure*}[t]
    \centering
    \includegraphics[width=\textwidth]{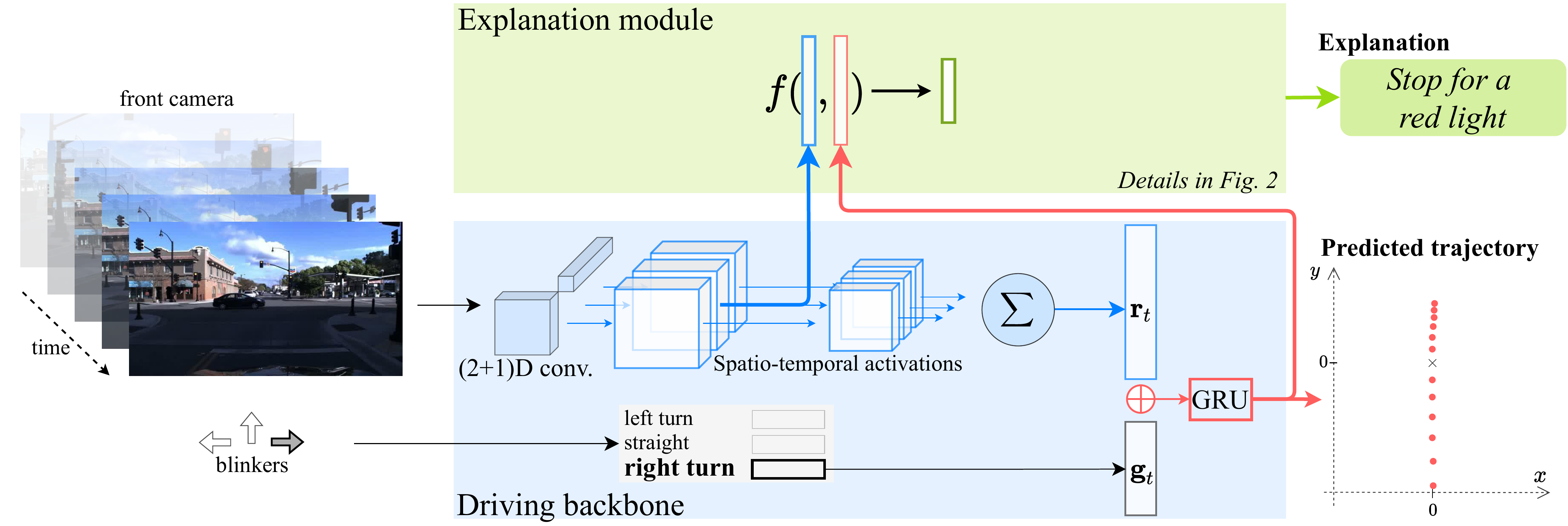}
    \vspace{-0.4cm}
    \caption{\textbf{Model overview.} The architecture is composed of an explanation module (green box) on top of a self-driving backbone (blue box). The driving system takes both the camera stream and the blinker state as input to predict the future trajectory of the car from a bird-eye view. Simultaneously, the explanation module produces a justification for the driving decision by jointly processing the high-level driving decision and intermediate perceptual features extracted from the driving system.
    \vspace{-0.2cm}
    }
    \label{fig:overview}
\end{figure*}

% EXPE
% we make experiments to show that we are generic and that our approach works
Our design choices are validated on two datasets, in different settings, thus showing the generic aspect of our approach.
% HDD: classes, online
First, on the Honda Deep Drive (HDD) dataset \cite{hdd_cvpr2018}, which associates each video frame to a \textit{cause} label, explanations in the form of cause labels are produced as the system drives, \ie in an online fashion.
% BDDX: language, offline
Second, with the Berkely Deep Drive eXplanation (BDD-X) dataset \cite{bddx}, explanations take the form of natural language justifications, generated in an offline fashion.
% Code available
%The code to reproduce our experiments will be made available.
%Besides, our approach to an end-to-end model, we adapt our explainer module to a setting where causes are expressed as natural language explanations 
%, produced in an offline fashion on the Berkeley Deep Drive eXplanation (BDD-X) dataset \cite{bddx}.

The contribution is three-fold: (1) we equip self-driving systems with a module providing introspective explanations, (2) we draw a comparison with VQA systems and adapt the BLOCK module to fuse high-level driving decisions with intermediate perceptual features, and (3) we exhibit the versatility of our approach 
% our claims are backed}
with experiments in various settings (online/offline, classification/generation). %exhibiting the versatility of our approach.

\section{Architecture overview}
\vspace{-0.1cm}

\label{sec:model}
In this section, we present \acro{}, a module providing explanations for a driving system's decision, as illustrated in Figure~\ref{fig:overview}.
%As illustrated in Figure~\ref{fig:overview}, the general architecture is composed of a driving system as the backbone and an \textit{explanation module} that gives high-level explanations as backbone drives.
% Online
Explanations are provided in an \emph{online} fashion as only the current and previous inputs are used. 
% Correlation
The central design choice of \acro{} is that decisions are explained by analyzing the correlations between mid-level perception and high-level decision signals.
This corresponds to \textit{introspective explanations} as the network accesses the driving backbone's internal representations.
The architecture's backbone is presented in Section~\ref{subsection:backbone_overview}, and the explanation module in Section~\ref{subsection:model:explainer}.
%We first detail the architecture's backbone in \autoref{subsection:backbone_overview} and we then present the explanation module in details in \autoref{subsection:model:explainer}.

\subsection{Driving backbone}
\vspace{-0.1cm}

\label{subsection:backbone_overview}
As in previous work \cite{chauffeurnet,infer,neural_motion_planner,desire,precog}, the driving system is a vision-based end-to-mid trajectory prediction model: at each timestep $t$, our backbone network predicts the future positions the ego-vehicle should reach.
%This approach is denoted \textit{end-to-mid}, as opposed to \textit{end-to-end} driving systems that directly predict vehicle controls.
%A predicted trajectory conveys richer information than a predicted vehicle control, and it can be directly fed to an actuator control system \cite{DBLP:conf/iccv/ChenSKX15,DBLP:conf/ivs/Al-QizwiniBAR17}.
It takes as input the current frame $\tens{I}_t$ and the $T$ previous frames $\tens{C}_t = [\tens{I}_{t-T}, ..., \tens{I}_{t-1}]$ of the front camera.
These $(T+1)$ color images are stacked to form $\tens{X}_t = [\tens{I}_{t-T}, ..., \tens{I}_{t-1}, \tens{I}_{t}]$, represented at the left of Figure~\ref{fig:overview}. 
%This input tensor is passed to a 3D Convolutional Neural Network,  illustrated in blue in \autoref{fig:overview}:
We pass this input tensor to a 3D Convolutional Neural Network to obtain $\mat{r}_t = \dcnn \left( \tens{X}_t \right) \in \mathbb{R}^d$, the vector representing the input image sequence $\tens{X}_t$. % as illustrated in blue in \autoref{fig:overview}:
%\begin{equation}
%     \mat{f}_t = \dcnn \left( \tens{X}_t \right),
%\end{equation}
%where $\mat{f}_t \in \mathbb{R}^d$ is a vector representing the input image sequence $\tens{X}_t$.
%\eloi{maybe not "we choose"? rather make it look like a standard choice? use a ref?}
Note that our setting is generic enough to be used with any driving architecture instead of our $\dcnn$, provided that intermediate representations can be accessed. 
Here, we experimented with an R(2+1)D network \cite{tran2018closer}. It processes input data through a series of five residual convolutional blocks, where 3D convolution kernels are factorized into separate spatial and temporal components.
%Overall, the driving module identifies spatio-temporal patterns at multiple hierarchical levels by learning a representation of the scene that is relevant to the prediction of driving decisions.

%\begin{figure}[t]
%    \centering
%    \includegraphics[width=\textwidth]{figures/driver.pdf}
%    \caption{\textbf{Driving module.} Our self-driving system is based on the R(2+1)D architecture \cite{tran2018closer} for video processing. A series of factorized 3D convolutions is applied to the streaming video frames $[\tens{I}_{t-T}, ..., \tens{I}_{t-1}, \tens{I}_t]$. At the last layer, spatio-temporal activations are aggregated to form $\mat{f}_t$. This vector is concatenated with a high-level goal signal $\mat{g}_t$ and passed to a $\gru$ network that predicts future points $\left[\left(\hat{x}_{t,k}, \hat{y}_{t,k}\right) \right]_{k=1..K}$.}
%    \label{fig:driver_rnn}
%\end{figure}

% Justification for the goal
%Importantly, the task of driving a car can hardly be seen as a simple sequence of reactions to perceptual signals.
To predict the future trajectory of the vehicle, the model needs to be informed about high-level goals of the car. %is required to achieve.
We represent the input goal as a categorical variable $g_t \in \left\{ \textit{left}, \textit{straight}, \textit{right} \right\}$ that indicates the state of the car turn signals, as in previous works \cite{codevilla2018end}. % activated by the human driver that our model tries to mimic.
The goal $g_t$ is then associated to a vector $\mat{g}_t$ of trainable parameters through an embedding table, shown in the bottom of Figure~\ref{fig:overview}.

% RNN version
In our framework, the future vehicle positions are generated by a recurrent model, here a $\gru$ network \cite{cho2014learning}, similarly to previous work \cite{desire}.
At every time step $t$, $K$ future positions are predicted by the driving system: for each decoding step $k \in \{1\dots K\}$, the network takes the same representation $\left[\mat{r}_t \oplus \mat{g}_t \right]$ as input and updates its internal state $\mat{h}_k$ to predict a future position $(\hat{x}_{t,k}, \hat{y}_{t,k})$:
\vspace{-0.05cm}
\begin{equation}
    \big( \left( \hat{x}_{t,k}, \hat{y}_{t,k} \right), \mat{h}_k \big) = \gru \left( \left[ \mat{r}_t \oplus \mat{g}_t \right], \mat{h}_{k-1} \right),
\end{equation}
where $\oplus$ designates the concatenation operator.

The first internal state $\mat{h}_0$ is initialized with the null vector.
%To summarize, at every time step $t$, $K$ future positions are predicted by the driving module.
% Training
The driving backbone is trained under the imitation learning paradigm with a mean squared error regression objective $\tens{L}_\textit{drive}$. % with : 
%\begin{equation}
%\label{eq:driver_loss}
%    \tens{L}_\textit{drive} = \frac{1}{T} \sum_{t=1}^T \sum_{k=1}^K \sqrt{ \left( x_{t, k} - \hat{x}_{t,k}\right)^2 + \left( y_{t, k} - \hat{y}_{t,k}\right)^2},
%\end{equation}
%where $x_{t,k}$ and $y_{t,k}$ are the ground-truth lateral and longitudinal coordinates $k$ steps ahead when the ego-vehicle is at time $t$.
%These coordinates are defined relatively to the current position of the vehicle: $x_{t,0} = y_{t,0} = 0$.
As we see in Section~\ref{subsubsection:experiments:bddx:newsetting}, this \textit{end-to-mid} architecture can be modified to fit into the \textit{end-to-end} setting. To do so, we replace the recurrent prediction head by a regression, trained to directly fit the vehicle controls.

\subsection{Explanation module}
\vspace{-0.1cm}
\label{subsection:model:explainer}
% On propose un explainer
As humans cannot reasonably decipher and interpret the driving system's inner processing, we build an explanation module to justify the driving decisions.
% On propose un explainer qui mélange
This module is designed to reason about internal representations of the driving backbone, under the light of the driving decision.
% Claim: il faut melanger
Indeed, to explain why a specific trajectory was output by the driver, we posit that merging intermediate spatio-temporal perceptual features with the higher-level decision features is key. 
% car plusieurs cause donne meme comportement donc il faut remonter dans les features
As exemplified in the introduction, various causes may collapse to produce a same output: a car braking could be caused by multiple events such as a red light, the presence of a pedestrian or a stop sign.
% Donc il faut a la fois les features et a la fois la decision
Hence, for the system to accurately recognize that the vehicle stops because of a red light, it should be given the information that the ego-vehicle chose to stop (\textit{decision}), but also that there is a red light (\textit{perception}).

\begin{figure*}[t]
    \centering
    \includegraphics[width=0.99\textwidth]{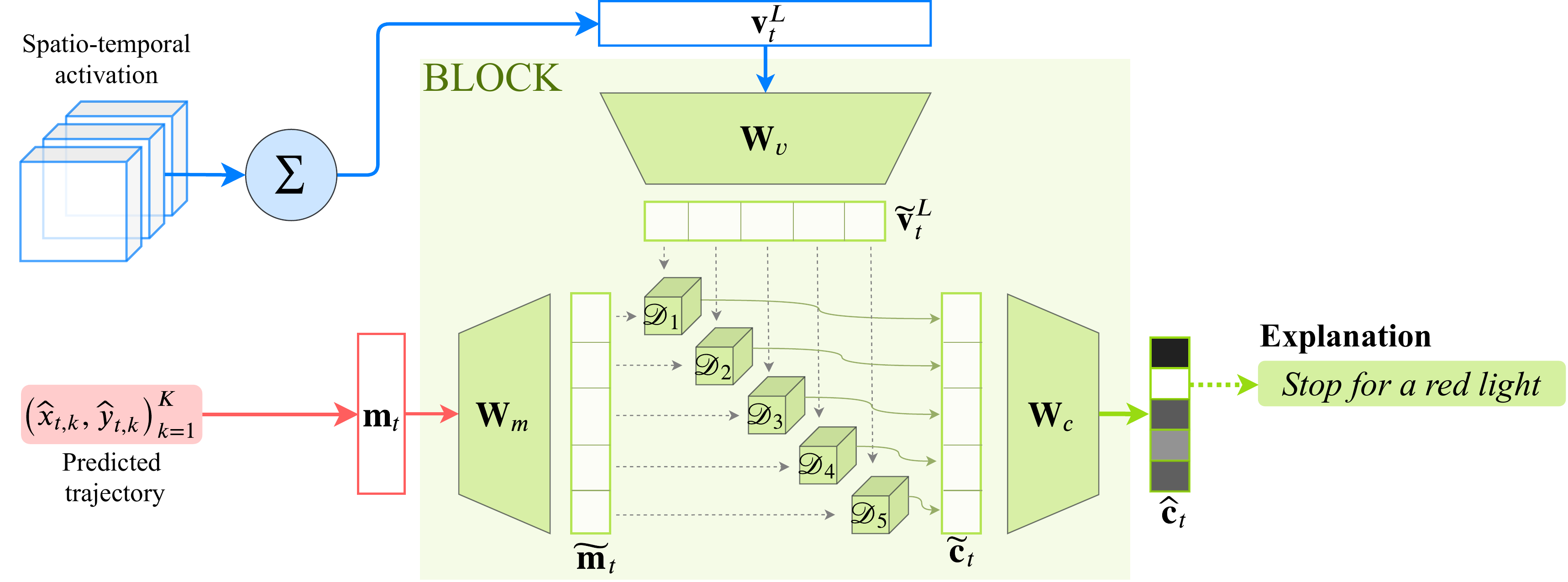}
    %\vspace{-0.1cm}
    \vspace{-0.4cm}
    \caption{\textbf{Explanation module.} The structure of BLOCK bilinear fusion enables to find relevant correlations between a perceptual mid-level representation $\mat{v}_t^L$ and high-level decision features $\mat{m}_t$. This forms $\mat{\hat{c}}_t$ which scores each possible explanation.
    \vspace{-0.3cm}
    }
    \label{fig:explainer}
\end{figure*}

\subsubsection{Overview}
\vspace{-0.1cm}

% Overview and notations
For a given residual block $L$ of the $\dcnn$, we extract spatio-temporal activation maps $\tens{V}_t^L \in \mathbb{R}^{t_L \times h_L \times w_L \times d_L}$, which contain localized information about the input frame sequence.
We aggregate all this information using a space-time average pooling to obtain $\mat{v}_t^L$; this vector contains intermediate perceptual features.
Besides, we define the vector $\mat{m}_t~=~\left[ \hat{x}_{t,1}, \hat{y}_{t,1}, ... , \hat{x}_{t,K}, \hat{y}_{t,K} \right]$, the flattened vector coordinates of all future positions predicted by the recurrent network.
Both the decision vector $\mat{m}_t$ and the perceptual features $\mat{v}_t^L$ are fused together with a function $f$ to form $\mat{\hat{c}}_t = f(\mat{m}_t, \mat{v}_t^L)$, where $\mat{\hat{c}}_t \in \mathbb{R}^{d_c}$ is the predicted probability distribution over each of the $d_c$ candidate causes.

We now discuss the instantiation of the fusion operation $f$.
Its design is not trivial as it merges information from vectors of a very different nature: $\mat{v}_t^L$ contains perceptual features extracted from a deep network and $\mat{m}_t$ is a list of 2D coordinates.
We thus leverage the family of bilinear models for multi-modal fusion \cite{benyounescadene2017mutan,DBLP:conf/iclr/KimOLKHZ17,mfb,ben2019block} 
as these fusion techniques have proven to be effective in contexts with heterogeneous media as for Visual Question Answering.
In their general form, bilinear models are parameterized by a tensor of trainable weights $\tens{T} \in \mathbb{R}^{2K \times d_L \times d_c}$, and defined by:
\begin{equation}
    \label{eq:bilinear}
    \mat{\hat{c}}_t = \tens{T} \times_1 \mat{m}_t \times_2 \mat{v}_t^L,
\end{equation}
where $\times_i$ refers to the mode-$i$ product. In other terms, each output dimension can be written as $\mat{\hat{c}}_t[i]~=~\mat{m}_t^\top~\tens{T}_i~\mat{v}_t^L$, where $\tens{T}_i~\in~\mathbb{R}^{2K \times d_L}$ is the $i$-th slice of $\tens{T}$.

As we need to model rich correlations between mid- and top-level features of the network, we use the BLOCK fusion \cite{ben2019block}, a recent and efficient bilinear variant, where $\tens{T}$ is factorized by constraining the \textit{block-term ranks} of the tensor. Formally, this model specifies a structure on $\tens{T}$ such that Equation~\ref{eq:bilinear} can be expressed a sequence of light operations, as illustrated in Figure~\ref{fig:explainer}:
\begin{align}
    \mat{\tilde{m}}_t &= \mat{W}_m \mat{m}_t,\hspace{0.1cm}\quad\quad\quad\quad\quad %\\
    \mat{\tilde{v}}_t^L = \mat{W}_v \mat{v}_t^L \\
    \mat{\tilde{c}}_t &= \tens{D} \times_1 \mat{\tilde{m}}_t \times_2 \mat{\tilde{v}}_t^L, \quad\quad%\\
    \mat{\hat{c}}_t = \mat{W}_c \mat{\tilde{c}}_t,
\end{align}
where $\tens{D}$ is a block-diagonal tensor.

Finally, the output vector is transformed into a probability distribution $\mat{p}_t$ over possible classes using the softmax operator: $\mat{p}_t=\mathrm{softmax}(\mat{\hat{c}}_t)$. In Section~\ref{subsubsection:experiments:bddx:newsetting}, we extend this classification setup to generate \textit{natural language explanations}, by using a language model instead of the classification layer.

\subsubsection{Learning}
\vspace{-0.1cm}
% LEARNING, HYPOTHESIS AND EXPERIMENTS
% We would like to have driving session data with annotations
Training the explanation module is in itself a challenge. 
To meet the standard supervised learning setting, ground-truth explanations for the driving system are needed. 
In other words, the training algorithm should be given annotations about why this specific driving model is taking its decision.
% But it does not exist
As we do not have access to this type of annotations, we make the following hypothesis: 
% Then we make an hypothesis
when a self-driving system is trained to mimic a human driver, explanations for the neural network's decisions coincide with justifications of the human driver.
% Why this hypothesis is important and what it enables us to do
This assumption allows us to leverage \textit{human driver justifications} as a proxy to ground-truth explanations of the driving model.
%\eloih{ TODO The mechanisms by which we human take actions and those through which we justify them may be different, as argued by \citeauthor{mythos_interpretability} \shortcite{mythos_interpretability}. }

Recent work of \cite{hdd_cvpr2018,li2020learning} propose to recognize action justifications (such as ``stop for red light'' or ``deviate for parked car'') in driving videos.
While they tackle the task of identifying causes of driving behaviors, these works are different from ours as they do not involve any \emph{driving model}. 
From a practical point-of-view, this means that when predicting a cause, we never use input images directly and we only allow ourselves to look at internal representations of a driving backbone.

Under the supervision of the ground-truth cause $c$, the explanation module is optimized with: % the cross-entropy:
\vspace{-0.1cm}
\begin{equation}
    \tens{L}_\textit{explain} = - \frac{1}{T} {\textstyle\sum}_{t=1}^T \log \mat{p}_t[c].
\end{equation}
Objectives $\tens{L}_\textit{drive}$ and $\tens{L}_\textit{explain}$ are linearly combined and jointly optimized.

\subsection{Link with Visual Question Answering (VQA)}
\vspace{-0.1cm}
The BLOCK fusion employed here was originally designed for VQA systems, where the representations of the question and the image are fused together to output the correct answer. 
VQA systems require ways to model interactions between feature vectors. Some work perform this multi-modal fusion using bilinear models, which are made tractable by sampling methods like MCB \cite{mcb} or through tensor factorization such as MUTAN \cite{benyounescadene2017mutan}, MLB \cite{DBLP:conf/iclr/KimOLKHZ17}, MFB \cite{mfb} or BLOCK \cite{ben2019block}.
In other work of \cite{perez2018film}, the FiLM layer is designed to perform fusion by linearly modulating visual features, conditioned on language vectors. Even if these fusion techniques were originally developed for VQA, many other applications benefit from those methods. Indeed, fusing multiple representations in deep learning architectures has drawn interest in visual relationship detection \cite{ben2019block}, image generation with StyleGAN \cite{Karras_2019_CVPR} or conditional domain adaptation \cite{long2018conditional}.

We can draw a parallel between fusion-based VQA systems and our explanation module. In our case, a question in the form of a trajectory is asked, and the explanation module builds the answer from the driving backbone's representations. More generally, our work illustrates the power of tensor-based methods to model fine-grained correlations between heterogeneous high-dimensional features.
\vspace{-0.2cm}
% Broadly, the explanation system answers the question ``Why did the driving system chose \textit{this} trajectory?'', regarding the internal state of the driving backbone. 

%\hedi{Point sur la fusion/VQA} \eloi{c'est à dire ?}
%
%\hedih{
%\begin{itemize}
%    \item La fusion utilisée ici a été initialement développée pour le vqa
%    \item On peut voir notre explanation model comme un système de vqa
%    \item Question = trajectoire prédite, et image = état interne du systeme. 
%    \item De façon générale, on pense que les factorizations de tenseurs sont de bonnes façons de modéliser des corrélations entre des variables vectorielles hétérogènes de hautes dimensions (ici une trajectoire vs. état interne d'un système)
%\end{itemize}
%}

%\hedi{Point sur le training: hypothèse d'équivalence entre les justifications humaines et les explications du driving model. On l'a déjà mentionné en intro, est-ce qu'on en remet une couche ?}
%\eloi{Ou alors on la met que ici et on évite l'embrouille en intro ?}

\section{Experiments}
\label{sec:experiments}
\vspace{-0.1cm}
We validate the design of \acro{} in Section~\ref{subsection:experiments:hdd}. In Section~\ref{subsection:experiments:bddx}, we show that \acro{} can be adapted to different settings, with an extension to offline natural language explanations.

\subsection{Online introspective explanations}
\vspace{-0.1cm}
%\subsection{Evaluation of online introspective explanations}
\label{subsection:experiments:hdd}
This section presents an experimental study of the explanation module presented in Section~\ref{sec:model}. 
We evaluate the quality of the explanations provided by \acro{} with respect to previous action recognition works. The design of the explanation module is validated and we show (1) the effectiveness of using both perceptual and high-level features, and, (2) the relevance of employing a BLOCK fusion. 

% \begin{table}[]
%     \centering
%     \caption{Cause prediction}
%     \begin{tabular}{@{}lccccccc@{}}
%     \toprule
%     \noalign{\smallskip}
%     System & mAP & Congestion & Sign & \makecell{Red \\light} & \makecell{Crossing \\ vehicle} & \makecell{Parked \\ vehicle} & \makecell{Crossing \\ pedestrian} \\
%     \noalign{\smallskip}
%     \hline
%     \multicolumn{8}{c}{Online models} \\
%     \hline
%     CNN+Sensors \cite{hdd_cvpr2018} & 28.25 & 39.72 & 46.83 & 45.31 & - & 7.24 & 2.15 \\
%     R(2+1)D baseline (ours) & 49.51 & 79.34 & 67.3 & 76.87 & 28.63 & 10.42 & 34.47 \\
%     Multi-task E (ours) & 49.22 & 84.2 & 71.07 & 78.20 & 32.78 & 12.67 & 16.41 \\
%     Multi-task B (ours) \\
%     Multi-task bilinear (ours) & \textbf{51.04} & 85.21 & 67.79 & 79.53 & 30.65 & 16.33 & 29.77 \\
%     \hline
%     \multicolumn{8}{c}{Offline models} \\
%     \hline
%     I3D \cite{li2020learning} & 43.9 & 64.8 & 71.7 & 63.6 & 21.5 & 15.8 & 26.2 \\
%     I3D+GCN \cite{li2020learning} & 49.9 & 74.1 & 72.4 & 76.3 & 26.9 & 20.4 & 29.0\\
%     R(2+1)D baseline (ours) & 50.47 & 85.07 & 74.01 & 80.26 & 28.99 & 12.84 & 21.64 \\
%     Multi-task (ours) & \textbf{51.72} & 76.9 & 71.62 & 82.41 & 32.74 & 20.36 & 27.03 \\
%     \bottomrule
%     \end{tabular}
%     \label{tab:cause_prediction}
% \end{table}

%\setlength{\tabcolsep}{1.5pt}

\newcommand{\fz}[1]{\fontsize{7}{0}{\selectfont #1}}
\newcommand{\ffz}[1]{\fontsize{8}{0}{\selectfont #1}}

\begin{table*}[t]
    \caption{\textbf{Comparison to state-of-the-art on the \textit{cause} labels of \hdd{}.} % Models are trained on \hdd{} train split and the results are computed on the test split.
    Results in bold are the best among \textit{online} models. For the sake of completeness, we also report performance of offline models.}
    %\resizebox{\textwidth}{!}{%
    \label{tab:cause_prediction}
    \centering
    \vspace{-0.2cm}
    \begin{tabular}{@{}r@{\hspace{0.2cm}}c@{\hspace{0.3cm}}ccccccc@{\hspace{0.35cm}}c@{}}
    \toprule
    %\noalign{\smallskip}
    \noalign{\vspace{-0.09cm}}
    & & \multicolumn{6}{c}{\ffz{Individual causes}} & \\
    \cline{3-8}
    System & \makecell[b]{\fz{Online}/\\\fz{Offline}} & \fz{Congest.} & \fz{Sign} & \makecell[b]{\fz{Red} \\\fz{light}} & \makecell[b]{\fz{Crossing} \\ \fz{vehicle}} & \makecell[b]{\fz{Parked} \\ \fz{vehicle}} & \makecell[b]{\fz{Crossing} \\ \fz{pedestrian}} & \makecell[b]{Overall \\ mAP} & \makecell[b]{Driver \\ MSE} \\
    %\noalign{\smallskip}
    \hline
    %\noalign{\smallskip}
    & & \multicolumn{7}{c}{\textit{Action recognition (no driver)}} \\
    \noalign{\smallskip}
    %\hline
    %\noalign{\smallskip}
    CNN+Sens. \cite{hdd_cvpr2018} & On. & 39.72 & 46.83 & 45.31 & --- & 7.24 & 2.15 & 28.25 & $\times$ \\
    I3D \cite{li2020learning} & Off. & 64.8 & 71.7 & 63.6 & 21.5 & 15.8 & 26.2 & 43.9 & $\times$ \\
    I3D+GCN \cite{li2020learning} & Off. & 74.1 & 72.4 & 76.3 & 26.9 & 20.4 & 29.0 & 49.9 & $\times$ \\
    %\noalign{\smallskip}
    \hline
    %\noalign{\smallskip}
    & & \multicolumn{7}{c}{\textit{Driver only (no explanation)}} \\
    %\noalign{\smallskip}
    %\hline
    %\noalign{\smallskip}
    Driver & On. & $\times$ & $\times$ & $\times$ & $\times$ & $\times$ & $\times$ & $\times$ & \textbf{1.33} \\
    %\noalign{\smallskip}
    \hline
    %\noalign{\smallskip}
    & & \multicolumn{7}{c}{\textit{Introspective explanation}} \\
    \noalign{\smallskip}
    %\hline
    %\noalign{\smallskip}
    Multi-head & On. & \textbf{81.25} & \textbf{66.59} & 75.46 & 31.21 & 10.24 & 25.62 & 48.39 & 1.36 \\ 
    % Multi-head + blinker & On. & 72.56 & \textbf{66.68} & \textbf{83.22} & 38.38 & \textbf{14.88} & 24.15 & 49.98 & 1.39 \\
    \acro{} & On. & 80.38 & 63.41 & \textbf{81.94} & \textbf{41.19} & \textbf{12.18} & \textbf{27.19} & \textbf{50.96} & \textbf{1.33} \\ 
    %\acro{} no blinker & \textbf{82.29} & 63.40 & 79.47 & 38.48 & 12.01 & 16.77 & 48.74 & 1.39 \\
    % XDrive 5 & Online & 81.38 & 66.16 & 82.37 & 36.06 & 9.08 & 24.13 & 1.36 & 49.86 \\
    % XDrive 4 & Online & 83.53 & 64.88 & 81.94 & 41.29 & 9.75 & 20.1 & 1.38 & 50.25 \\ 
    % XDrive 2 & Online & 86.47 & 55.34 & 70.35 & 30.77 & 6.68 & 8.92 & 1.44 & 43.09 \\
    \bottomrule
    \end{tabular}
    \vspace{-0.2cm}
    %}
\end{table*}

% \begin{table}[]
%     \centering
%     \caption{Cause prediction. Our models are trained on }
%     \begin{tabular}{@{}lcccccccc@{}}
%     \toprule
%     \noalign{\smallskip}
%     System & \makecell{Online/\\Offline} & mAP & Congestion & Sign & \makecell{Red \\light} & \makecell{Crossing \\ vehicle} & \makecell{Parked \\ vehicle} & \makecell{Crossing \\ pedestrian} \\
%     \noalign{\smallskip}
%     \hline
%     \multicolumn{9}{c}{Trained on \textit{train} split} \\
%     \hline
%     CNN+Sensors \cite{hdd_cvpr2018} & Online & 28.25 & 39.72 & 46.83 & 45.31 & - & 7.24 & 2.15 \\
%     I3D \cite{li2020learning} & Offline & 43.9 & 64.8 & 71.7 & 63.6 & 21.5 & 15.8 & 26.2 \\
%     I3D+GCN \cite{li2020learning} & Offline & 49.9 & 74.1 & 72.4 & 76.3 & 26.9 & 20.4 & 29.0 \\
%     \hline
%     \multicolumn{9}{c}{Trained on \textit{train*} split} \\
%     \hline
%     Baseline (ours) & Online & 48.05 & 74.75 & 63.24 & 79.43 & 27.19 & 13.34 & 30.63 \\
%     Bilinear model (ours) & Online & 49.55 & 79.10 & 62.58 & 79.89 & 30.82 & 15.33 & 29.56 \\
%     Baseline fixloss \\
%     ~~ bestmAP & Online & 47.21 & 82.70 & 65.35 & 77.0 & 25.66 & 13.18 & 19.39 \\
%     ~~ bestloss & Online & 46.25 \\
%     \bottomrule
%     \end{tabular}
%     \label{tab:cause_prediction}
% \end{table}

\subsubsection{HDD dataset \citep{hdd_cvpr2018}}
\vspace{-0.1cm}
%\eloi{Is it clear at this point that we are online / classification / end-to-mid?}
%\paragraph{HDD dataset.}
To study driving behavior causes from a driver-centric view, the Honda Deep Drive (\hdd{}) dataset has been introduced \cite{hdd_cvpr2018}.
%The dataset gathers data from front camera, GPS, \lidar{} and CAN-bus sensors, and annotations about driving decision causes such as ``stop for red light'' or ``deviate for parked car''. 
%goal
%It also contains annotations about the driver's goal (\eg ``intersection passing'', ``left lane branch''), for which some driving action recognition systems have been developed \cite{xu2019temporal,li2020learning,DBLP:journals/corr/abs-1910-00628}.
The dataset gathers 137 driving sessions recorded in the San-Francisco Bay Area. In total, 104 hours of human driving videos have been acquired, synchronized with CAN bus signals and GPS/IMU information. We only use front camera images, turn signals and GPS positions to train our model. Additionally to these signals, several layers of frame-level annotations are provided to describe the driving behavior. Among these layers, we focus on the \textit{cause} labels which convey information about the underlying reason explaining the driver's behavior (\eg ``stop for red light'' or ``deviate for parked car''). Following the literature on action detection \cite{DBLP:conf/cvpr/ShouCZMC17,hdd_cvpr2018}, we evaluate each frame independently and we report the value of the mean Average Precision (mAP).

%\textbf{Implementation details} are given in the supplemental of the paper.

\subsubsection{State-of-the-art comparison}
\vspace{-0.1cm}
\label{subsubsection:experiments:hdd:sota}
%\eloi{check the claims below with respect to the new introduction}
%\eloi{do we really want to tell our work apart from previous action recognition works on HDD?}
As stated in the introduction, our work is the first driving system equipped with an introspective explanation module.
Previous works using \hdd{} would perform driving scene recognition, mainly on \textit{goal-oriented} labels.
Even if some of them are trained on the \textit{cause} labels, which explain the behavior of the human driver, they adopt a setup of action recognition and do not consider any driving model. 
To measure the relevance of our online introspective explanations, we compare against state-of-the-art methods for action recognition:
\begin{itemize}
\vspace{-0.1cm}
    \item \textbf{CNN+Sensors} \cite{hdd_cvpr2018} where convolutional features and sensor values are merged before being passed to an LSTM that models temporal dependencies;
\vspace{-0.05cm}
    \item \textbf{I3D} \cite{CarreiraZ17}, a fined-tuned I3D network implemented by \cite{li2020learning};
\vspace{-0.05cm}
    \item \textbf{I3D+GCN} \cite{li2020learning} where the scene is analyzed by multiple perception models that infer depth, detect objects and segment the relevant areas. This information is processed by multiple graph convolution networks in order to recognize the cause label.
\end{itemize}
Importantly, \textbf{I3D+GCN} \cite{li2020learning} is an \textit{offline} model as causes are predicted using previous and \emph{future} frames.
Besides, we also evaluate an internal baseline, named \textbf{Multi-head}, which simply adds an auxiliary branch to the last layer of the $\dcnn$ without using the predicted trajectory, nor any fusion.
Also, we note \textbf{Driver} the backbone driving system, \ie without the explanation module.
%\begin{itemize}
%    \item \textbf{Driver} which consists of only the driving module. 
%    \item \textbf{Multi-head} which simply adds an auxiliary branch to the last layer of the $\dcnn$ without using the predicted trajectory.
%\end{itemize}

%\begin{figure}[t]
%    \centering
%    \includegraphics[width=0.85\linewidth]{figures/layer_choice_mt.pdf}
%    %\vspace{-0.2cm}
%    \caption{\textbf{\acro{} performance by layer.} Blue points represent our explanation model, and the red dotted line is the multi-head baseline.}\label{fig:layer_choice_mt}
%\end{figure}

Results of this comparison are reported in Table~\ref{tab:cause_prediction}. First, we observe that \acro{} outperforms all action recognition systems, \emph{including offline ones}.
%BEEF provides explanations for decisions of a \emph{self-driving system} whereas [21] does cause recognition in \emph{human} driving videos. 
%Moreover, our model works in an \emph{online} fashion, a more challenging setting than the \emph{offline} way of [21].
%which is much more challenging the task tackled in [21] offline architecture, which means they output a cause label using \emph{future frames}, while
While the average performance of \acro{} is indeed higher than I3D+GCN, we notice a slight performance drop on some classes (\eg \textit{Sign}, \textit{Parked vehicle}).
% We hypothesize that this stems from the fact that these classes have a shorter time-span and this may benefit \emph{offline} models as they can access future frames, while these frames remain not accessible for BEEF. 
We hypothesize that this stems from the fact that annotations from these classes have a shorter time-span. This may benefit \emph{offline} models which can access future frames, while these frames remain not accessible for BEEF.
This is supported by the value $\rho=0.71$ of the Spearman correlation between the average duration of annotations from each class and the relative performance shift.
Overall, despite a slight drop in some classes, the online architecture of BEEF is suitable for real-world self-driving explanations, unlike the offline action recognition model of \citep{li2020learning}.

%\hedih{When comparing \acro{} with the best offline model of \cite{li2020learning}, it appears that our system is able to improve the results on some classes (\ie \textit{Red light} or \textit{Crossing vehicle}) and decreases them in others (\ie \textit{Sign} or \textit{Parked vehicle}). Given that our system works in an online fashion, its performance is expected to be lower than the offline systems of \citep{li2020learning}. Still, we want to understand the difference between those classes that make \acro{} work better or worst. Our hypothesis is that \emph{short sequences} are better recognized with offline systems, which are designed to anticipate the occurrence of an event. To support this idea, we measure the average duration of annotations for each class, as well as the relative difference between the precisions of \acro{} and I3D+GCN. The Spearman correlation between average duration and relative performance gain is 0.71, which intuitively means that the I3D+GCN performs better than \acro{} on short events. This stands in favor of our hypothesis.}

Besides, we see that \acro{} performs better than the Multi-head baseline, for which the explanation module only uses perceptual features.
This shows their complementarity with the predicted trajectory and stresses the importance of combining them together to provide explanations.
We note that by fusing the predicted trajectory with the perceptual features in \acro{}, the cause prediction may be indirectly influenced by the blinker signal used in the driving system.
% unlike in the Multi-head baseline.
To make sure that this does not solely explain the gain in predictive performance, we also evaluate a modified Multi-head model which uses the blinker information by concatenating $\mat{g}_t$ with the output of the $\dcnn$. 
This variant obtains an overall mAP of 49.98, a higher score than the original Multi-head but still below \acro{}.
This emphasizes the importance of the fusion module used in \acro{}, as both Multi-head architectures struggle to balance between %both
driving and explanation objectives. %, and have to sacrifice performances in each task.

\subsubsection{Choice of the layer $L$}
\begin{figure}[t]
\BottomFloatBoxes
\begin{floatrow}
\ffigbox
  {\includegraphics[width=\linewidth]{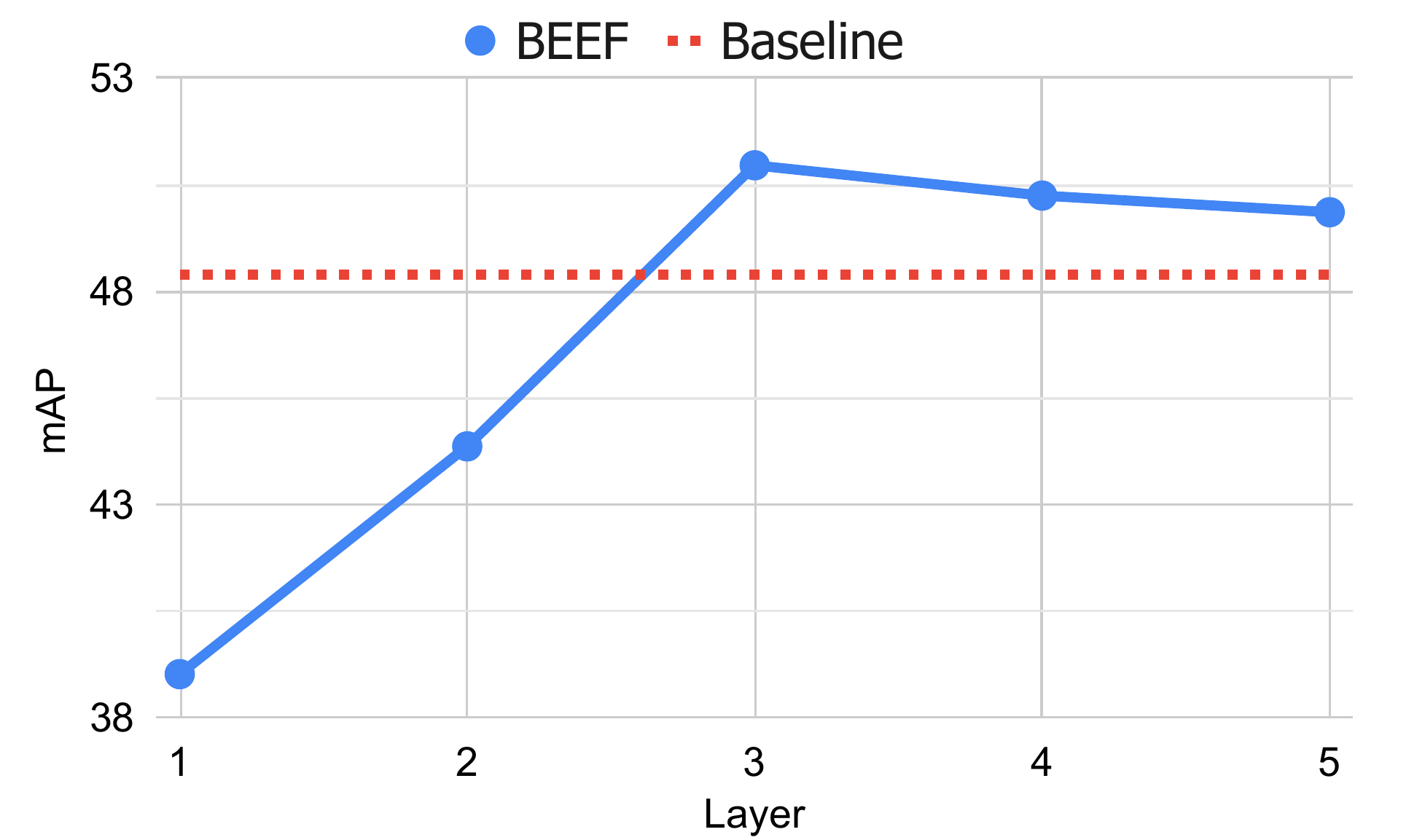}}
  {
  %\caption{\textbf{\acro{} performance by layer.} Blue points represent our explanation model, and the red dotted line is the multi-head baseline.}\label{fig:layer_choice_mt}
  \caption{\textbf{\acro{} performance by layer.}}\label{fig:layer_choice_mt}
  \vspace{-0.2cm}
  }
\killfloatstyle
\ttabbox
  {
  \setlength{\tabcolsep}{5pt}
%\begin{table}
%  %\setlength{\tabcolsep}{5pt}
  \begin{tabular}{@{}lcc@{}}
    \toprule
    Fusion model & mAP & Driver MSE \\
    \noalign{\smallskip}
    \hline
    Layer 3 & 45.39 & 1.38 \\
    Cat+MLP & 46.64 & 1.38 \\
    \hline
    MFB & 39.10 & 1.43 \\
    MLB & 38.59 & 1.37 \\
    MUTAN & 45.80 & 1.40 \\
    Bilinear & 49.82 & 1.42 \\
    BLOCK & \textbf{50.96} & \textbf{1.33} \\
    \bottomrule 
  \end{tabular}
  }
  %{\caption{\textbf{Fusion comparison.} Evaluation of different fusion models in the explanation module. Fusions are performed between driving decisions $\mat{m}_t$ and mid-level features $\mat{v}_t^3$.\label{tab:fusion_choice}\vspace{-0.2cm}}}
  {\caption{\textbf{Fusion comparison.} Fusions are performed between driving decisions $\mat{m}_t$ and mid-level features $\mat{v}_t^3$.\label{tab:fusion_choice}\vspace{-0.2cm}}}
  
%  \end{tabular}
%  %\vspace{0.15cm}
%  \caption{\textbf{Fusion comparison.} Evaluation of different fusion models in the explanation module. Fusions are performed between driving decisions $\mat{m}_t$ and mid-level features $\mat{v}_t^3$.\label{tab:fusion_choice}}
%\end{table}
  
%  \begin{table}
%     \centering
%     \caption{Comparing different fusion methods for a fusion between Prediction and Layer2. Trained on mini-train, results on mini-val}
%     \begin{tabular}{@{}lcc@{}}
%     %\hline
%     \toprule
%     Fusion model & mAP & Driver MSE \\
%     \hline
%     \bottomrule 
%     \end{tabular}
%     \label{tab:fusion_choice}
% \end{table}
\end{floatrow}
\vspace{-0.42cm}
\end{figure}

\vspace{-0.13cm}
\acro{} fuses the high-level decision vector $\mat{m}_t$ with low-level perception features $\mat{v}_t^L$. Here, we study how the performance of \acro{} depends on the chosen perception layer. Figure~\ref{fig:layer_choice_mt} shows the overall mAP cause prediction performances with respect to the layer used for the perceptual features in \acro{}. The horizontal dotted line corresponds to the \textbf{Multi-head} baseline shown in Table~\ref{tab:cause_prediction}.

Unsurprisingly, we observe that early layers poorly perform compared to the multi-head baseline. 
Early features of the network are too low-level to recognize relevant visual patterns for the cause identification. 
We observe a performance peak at the third layer. 
At this level, the representation contains rich perceptual signals, but still conveys enough cause information that is complementary to the final decision. 
As we use higher-level features of the network, we move away from the balance in the trade-off between representation richness and independence to the decision. In other words, the explanation module is fed with two vectors that convey similar information, which makes their combination less useful. 
Interestingly, we remark that \acro{} with layer 5 is better than the multi-head baseline: there is still information to extract from the interaction between the predicted trajectory and the hidden layers of the network.
This validates the design of the explanation module as a fusion between mid-level representations and high-level driving decisions.

\subsubsection{Fusion comparison}
\vspace{-0.13cm}
The design of the fusion operation used in the explanation module is a critical element of the architecture.
To take advantage of the complementarity between both inputs, the fusion operator must allow fine-grain correlations.
The experiment shown in Table~\ref{tab:fusion_choice}, compares the BLOCK fusion with other techniques, including:
\begin{itemize}
    \vspace{-0.1cm}
    \item \textbf{Cat+MLP} concatenates decision and mid-level features followed by a 2-layer perceptron,
    \vspace{-0.08cm}
    \item \textbf{MLB, MFB} and \textbf{MUTAN} are bilinear fusion methods developed for VQA \cite{DBLP:conf/iclr/KimOLKHZ17,mfb,benyounescadene2017mutan},
    \vspace{-0.08cm}
    \item \textbf{Bilinear} is the unstructured bilinear model that follows Equation~\ref{eq:bilinear}.
    \vspace{-0.08cm}
\end{itemize}

Moreover, we compare to a multi-head model with a 2-layer perceptron connected at layer 3, without any fusion. This model is referred to as \textbf{Layer 3}.

Using only Layer 3 performs poorly, which again supports the fact that driving decisions are important inputs for the explanation module.
Among all the compared fusions, BLOCK provides the most relevant explanations. 
It shows that this bilinear model is able, through its structure, to find the correlations between its inputs that carry relevant information for its task. 
Interestingly, BLOCK is the only fusion model that does not degrade the driving performance, while unstructured bilinear models heavily degrade the driving capacity. 
As this model does not involve mono-modal projections, it lacks flexibility and forces the trajectory prediction to adapt to the explanation task.

% \input{tables/blinker.tex}

%\input{tables/bdd/bdd_sample_short.tex}

%\subsection{Extension to natural language explanation generation}
\subsection{Offline natural language explanations}
\label{subsection:experiments:bddx}
\vspace{-0.12cm}

% Purpose of this subsection: dialog and showing that our approach is generic and flexible to other scenarios
Towards the long-term goal of developing human-vehicle dialogs, we aim at building a driving system that formulates explanations in \textit{natural language}.
We are motivated by the fact that open-domain sentences can convey finer and richer semantics than predefined classes.
%, which cannot capture the full diversity of possible explanations. %, as argued by \citeauthor{bddx} \shortcite{bddx}}.
% On adapte la sortie du module d'explication pour generer du langage
In this subsection, the cause classification becomes a language generation problem and we adapt the explanation module to enable the generation of textual explanations.
% On utilise BDD-X
To learn and evaluate our model, we use the recent BDD-X dataset \cite{bddx}, which provides natural language explanations for human driving sessions.

%\subsubsection{BDD-X dataset \cite{bddx}}
%\label{subsubsection:experiments:bddx:dataset}
%\vspace{-0.1cm}

\newcolumntype{L}{@{}>{\fontsize{7}{7}}l}
%\newlength{\columnspace}
%\setlength{\columnspace}{0.3cm}

\begin{table*}[t]
    \centering
    \begin{tabular}{@{}l r c L c L c L@{}}
    \toprule
    Task & Model & \multicolumn{2}{l}{\bleu} & \multicolumn{2}{l}{\meteor} & \multicolumn{2}{l}{\cider} \\
    \hline 
    \multirow{2}{*}{\makecell[l]{\textit{Action recognition} \\ {(no driver)}}} & {``because the light is red''} & 5.85 & & 10.74 & & 59.50 \\ % simple baseline
    & Vid-to-Text \cite{seqseqvideotext} & 6.33 & & 11.19 & & 53.35 \\ % reported from BDDX paper
    \hline
    \multirow{2}{*}{\makecell[l]{\textit{Rationalization} \\ {(offline)}}} & Rationalization \cite{bddx} & 6.52 & & 12.04 & & 61.99 \\  % reported from BDDX paper
     & Decision Features & 9.15&$\pm{}$.37 & 14.34&$\pm{}$.23 & 92.08&$\pm{}$1.3 \\ % workspace/logs/layer_bs32_lr0x0003_lstm64/output_seed_*
     \hline
    \multirow{4}{*}{\makecell[l]{\textit{Introspective} \\ \textit{explanation} \\ {(offline)}}} & SAA  \cite{bddx} & 7.07 & & 12.23 & & 66.09 & \\  % reported from BDDX paper
     & WAA \cite{bddx} & 7.28 & & 12.24 & & 69.52 & \\  % reported from BDDX paper
    & Layer 2 & 7.96 & & 13.51 & & 83.46 & \\  % workspace/logs/layer_bs32_lr0x0003_lstm64/layer2_seed_1234 
    & \acro{} & \textbf{9.81} &$\pm{}$.32 & \textbf{14.79}&$\pm{}$.22 & \textbf{97.31}&$\pm{}$2.3 \\ % logs/fusion_bs32_lr0x0003_lstm64/layer2_output_block_seed_*
    \bottomrule
    \end{tabular}
    \vspace{-0.1cm}
    \caption{\label{tab:bdd_caption}\textbf{Offline explanations generation on BDD-X.} Values are given in percentages, standard deviations are computed on three runs with different seeds.
    The first line corresponds to a baseline that always generates the same sentence. Results of the Vid-to-Text model are reported from \cite{bddx}.
    \vspace{-0.2cm}
    }
\end{table*}

\vspace{-0.1cm}
\paragraph{BDD-X dataset \cite{bddx}}
This dataset builds upon the BDDV dataset \cite{bddv} by adding \textit{explanations} to the driving sessions.
The dataset is composed of $6\,984$ videos, each lasting about 40 seconds and containing about 3.8 annotations on average. The dataset totals 77 video hours and 26K annotations.
Sensor data come from the front camera, GPS and IMU signals. 
In every driving session, open-domain natural language explanations are provided for each segmented sub-sequence. 

\subsubsection{Tailoring \acro{} to the new setting}
\vspace{-0.1cm}
\label{subsubsection:experiments:bddx:newsetting}
To fairly compare our approach with previous work on BDD-X \cite{bddx}, we need to make three modifications to the architecture presented in Section~\ref{sec:model}.

% Language
    \textbf{Natural language. \hspace{0.3cm}}
We adapt the explanation module to enable the generation of \textit{natural language} justifications, instead of explanation labels. In practice, we replace the classification layer by an autoregressive model instantiated with an attention-based LSTM network \cite{lstm} as done in \cite{bddx}.

% Driving: end-to-end
    \textbf{End-to-end backbone. \hspace{0.3cm}}
We change the driving module to be \textit{end-to-end}, instead of end-to-mid as presented in Section~\ref{sec:model}. In practice, the future trajectory prediction problem is simply replaced by a control command regression problem, and a single projection layer is used to output the acceleration and change of course values as in \cite{bddx}. As a consequence, we slightly change the source of high-level decision features in \acro{}. We use the last layer, \ie layer 5, instead of the predicted trajectory as done in Section~\ref{sec:model}. Indeed, contrary to end-to-mid models, end-to-end models output a weak signal about the future vehicle position by solely predicting the current acceleration and steering values.

% offline
    \textbf{Offline explanations. \hspace{0.3cm}}
We conduct the learning and evaluation phases in an \textit{offline} fashion as annotations are provided at the level of segmented driving sub-sequences.
%While our overall architecture is initially designed to work in the online setting, we can straightforwardly adapt it to the offline setting, by simply feeding the whole sequence as input and by generating a single justification for this sequence, as previously done on \bddx{} \cite{bddx}.
We straightforwardly adapt BEEF to the offline setting, by simply feeding the whole sequence as input and by generating a single justification for this sequence, as previously done on \bddx{} \cite{bddx}.

\begin{table*}%[b]
    \centering
    %A decoding temperature of $0.3$ has been used to encourage diversity.}
    \resizebox{\textwidth}{!}{%
    \begin{tabular}{@{}l l l l@{}}
        %\toprule
        \rotatebox{90}{\hspace{0.2cm} Extracted frame} & \includegraphics[height=30mm]{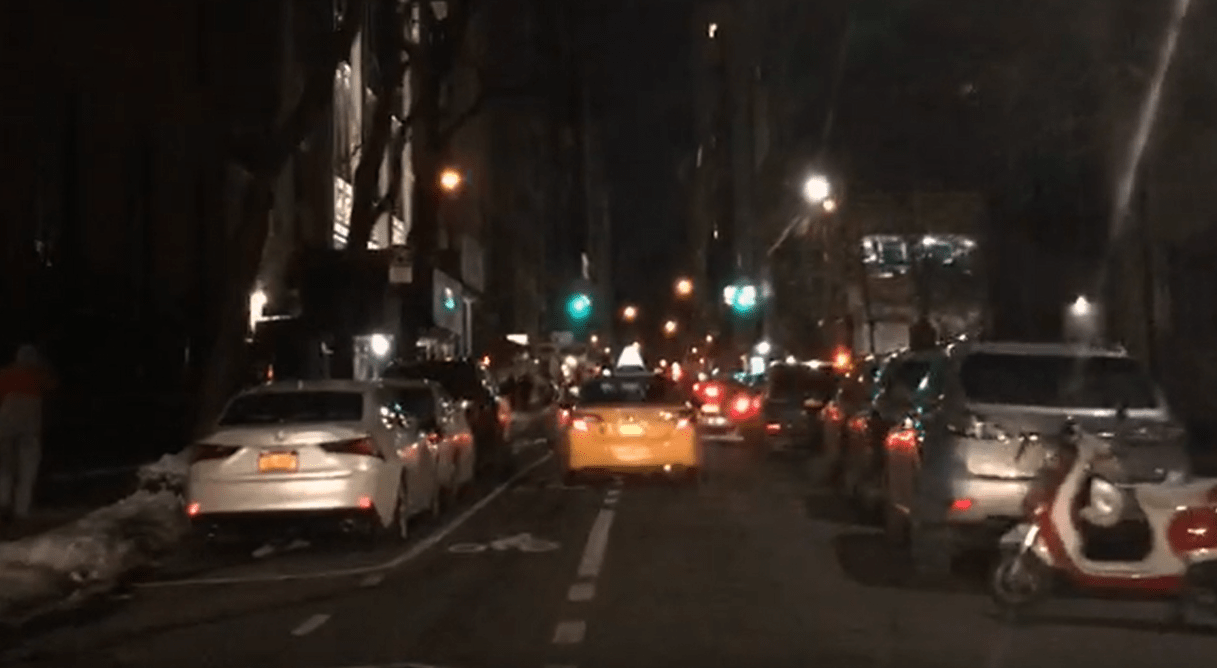}
        & \includegraphics[height=30mm]{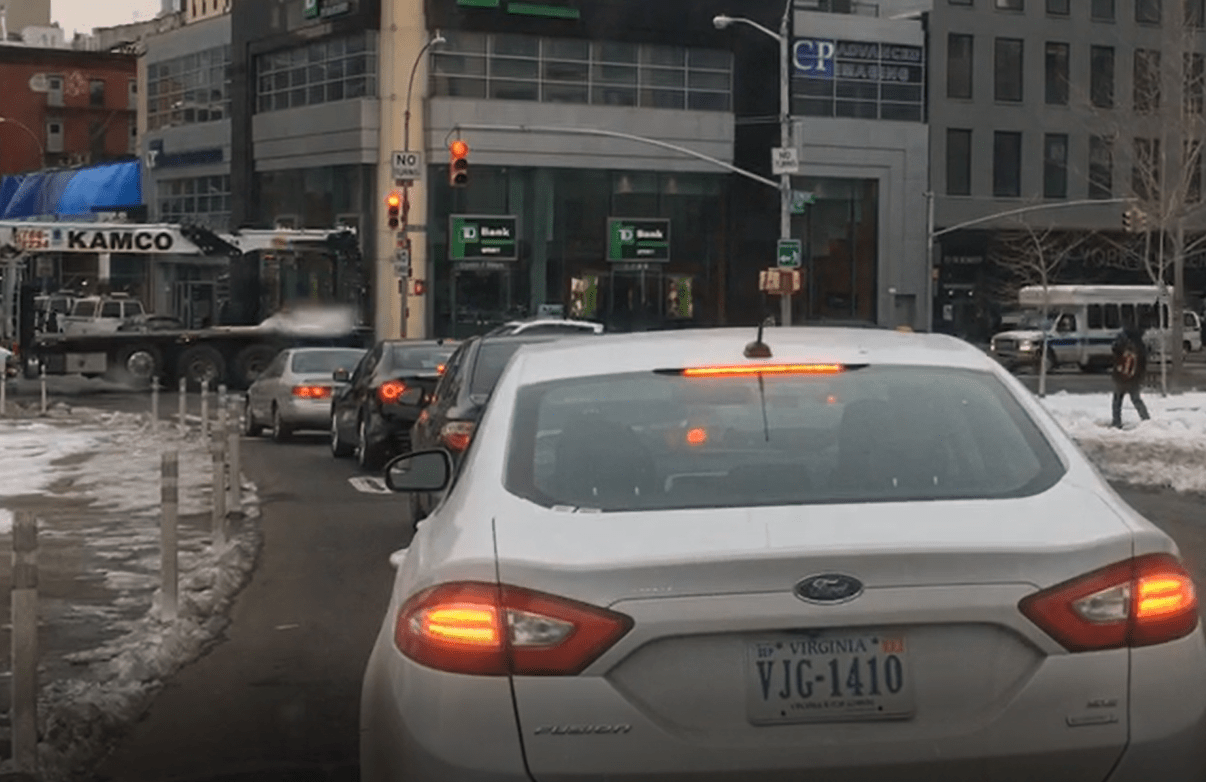}
        & \includegraphics[height=30mm]{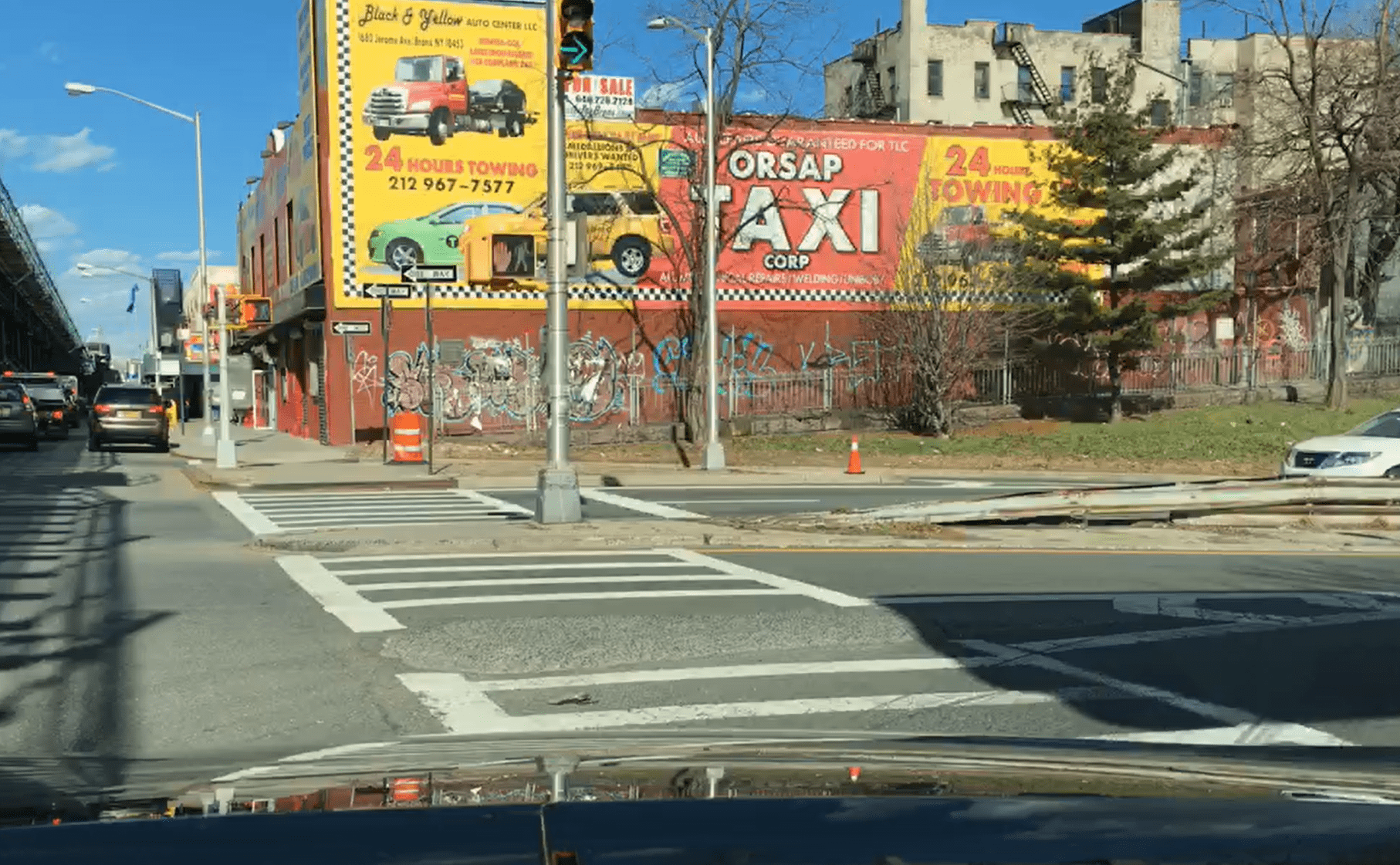} \\
        HL & because traffic is moving now & since the cars in front aren't moving & since the car is free to move right \\
        T=0 & because the light is green and traffic is moving & because the car in front has stopped & because the car is turning to the right \\
        T=0.3 & as the light turns green and traffic is moving & because the light is red & because the car is entering another street \\ % 1234
        T=0.3 & because the light is green and traffic is moving & as traffic ahead is stopped at a red light & to enter another road \\ % 2424
        %T=0.3 & because traffic is moving forward & because the car in front has stopped & because it s making a right turn \\ %  4242
        T=0.3 & because the light turns green & because traffic is stopped at a red light & because the road is clear of traffic \\ % 2512
        %T=0.3 & because the light turned green and traffic is moving & because traffic is stopped & to make a right turn  \\ % 2403
        %\bottomrule
    \end{tabular}
    \vspace{-0.25cm}
    \caption{\label{tab:bdd_temperature}\textbf{Samples of generated explanations}. HL stands for the Human gold Label. Other lines are generated by BEEF, with different runs obtained with various decoding temperature T: T=0 corresponds to the greedy decoding and lines with T=0.3 correspond to random decoding with T=0.3.
    %For illustration purpose, a frame has been extracted from each corresponding sub-sequence.
    \vspace{-0.55cm}
    }
    }
    \vspace{-0.25cm}
\end{table*}

\subsubsection{Competition, baselines and evaluation}
\vspace{-0.1cm}
\acro{} is compared against previous models on the BDD-X dataset.
This includes:
% action recognition
(1) a deep video-captioning model that does not involve any driver: \emph{Vid-to-Text} \cite{seqseqvideotext},
% rationalization
(2) an offline rationalization model, namely the \emph{Rationalization} baseline of \cite{bddx}, 
% introspective explanation
and,
(3) offline introspective explanation models, namely \emph{SAA} and \emph{WAA} \cite{bddx}, which constrain the attention of the explainer to align with the attention of the driving system either strongly or weakly respectively for SAA and WAA.
Moreover, as \acro{} fuses intermediate features from \textit{Layer 2} with the \textit{Decision features} of the last layer, we also evaluate against the two baselines considering these feature vectors independently.

%\subsubsection{Evaluation}
%\vspace{-0.1cm}
%\label{subsubsection:experiments:bddx:details}
Similarly to previous work \cite{bddx}, explanations are decoded in a greedy fashion and their quality is measured with the \bleu \cite{bleu}, \meteor \cite{meteor} and \cider \cite{cider} metrics. % Explanations are greedily generated by sampling the most likely word at each decoding step. %Implementation details are given in the supplemental.

\subsubsection{Results analysis}
\vspace{-0.1cm}
\label{subsubsection:experiments:bddx:results}

% Surpassing SOTA
Table~\ref{tab:bdd_caption} reports that \acro{} obtains leading results on \bddx{} by surpassing performances of previous works, namely \emph{Vid-to-Text}, \emph{SAA} and \emph{WAA}, by a large margin on all of the evaluation metrics. % (\bleu +2.5, \meteor +2.6, \cider +27.).
% Need for intermediate features
Importantly, \acro{} surpasses both the \emph{Layer 2} and \emph{Decision Features} baselines: % and this has two implications.
% In line with hdd and underlies the need for intermediate features
this is in line with results on \hdd{} presented in Section~\ref{subsubsection:experiments:hdd:sota} and it supports the claim that perceptual features (\eg Layer 2) and high-level decision features are complementary for introspective explanations;
% It also show that our approach is flexible
moreover, it shows the flexibility and adaptability of our approach to various settings whether end-to-mid or end-to-end, online or offline, cause classes or textual justifications.

% We do qualitative and we play on temperature to have diversity
Finally, we show some qualitative samples of explanations generated by \acro{} in Table~\ref{tab:bdd_temperature},  where the diversity of generated sentences is controled by tuning the temperature parameter of the decoding softmax.
% We observe diversity - Avantage du language - Avoir plusieurs modes
We observe reasonable and diverse explanations for various situations.
% Different generation are consistent with each other and can enrich senses
In particular, we note that the different justifications decoded with a temperature T=0.3 are consistent with one another and remark that obtained explanations are diverse in two ways: \emph{syntactically} and with respect to their \emph{completeness} level as explanations can be more or less exhaustive when multiple explanations are simultaneously valid (\eg the car stops for both a red light and stopped traffic).
% Limite, higher temperature: noise qui risque de changer la sémantique et non pas 
However, we empirically notice that when the decoding temperature is increased further (beyond 0.3), generated sentences start being \emph{semantically} undermined. 
Increasing completeness and syntactic diversity without impairing the underlying semantics constitutes an exciting future research direction.
%\hedih{At this point, they may not correspond to the true explanations anymore, and could even stop being consistent with one another. 
%Using the sampling temperature to increase syntactic diversity and completeness without decreasing the semantic correspondence to true explanations constitutes an exciting possible future work.}

% \eloi{finir note positive, ou future work en disant que c'est une piste de recherche intéressante de jouer sur la diversité syntactique et de complétude sans modifier la sémantique ?}
% TODO warning on persuasive explanations?
%Lastly, we warn 

%% Limite du language l'évaluation n'est pas easy
%\eloih{Herman [18] notes that we should be wary of evaluating interpretable systems using merely human evaluations of interpretability, because human evaluations imply a strong and specific bias towards simpler descriptions. He cautions that reliance on human evaluations can lead researchers to create persuasive systems rather than transparent systems. WARNING (in \cite{DBLP:conf/dsaa/GilpinBYBSK18})}
%% Température good pour la syntaxe ou donner des explication plus complètes, mais risque potentiel de changer la sémantique.
%% On risque de tomber dans les persuasive
%\eloih{TODO integrate below, and mix with temperature and sampling}
%As previously noted in \cite{DBLP:journals/corr/abs-1711-07414}, one has to be careful about human evaluation of interpretability: relying exclusively on human assessment may lead to \textit{persuasive} systems, sacrificing fidelity for user acceptance. 

\section{Conclusion}
\vspace{-0.1cm}
\label{sec:conclusion}
%Recall of main contributions
In this paper, we presented \acro{}, an architecture that provides explanations for the decisions taken by a driving system.
%the first driving model that has a ``verbose mode'', as it can give explanations for the decisions it takes while driving.
%
On top of a driving backbone,
%forwarded on a streaming driving video and predicting the future driving trajectory,
\acro{} produces introspective explanations by fusing together high-level decisions and intermediate perceptual features.
We showed that the BLOCK operator, originally developed to fuse multi-\textit{modal} inputs, can be efficiently leveraged to fuse multi-\textit{level} inputs.
%Leveraging recent research on bilinear fusion architectures for multi-modal inputs, we showed that 
%design the fusion with a BLOCK module.
Our approach is validated on huge real-world driving datasets, \hdd{} and \bddx{}, where the quality of explanations surpasses previous state of the art.
Besides, we showed the flexibility of \acro{} to various settings (online/offline, cause classes/natural language justifications). %, namely online or offline explanations, end-to-mid or end-to-end driving backbones, cause classes or richer natural language justifications. 
%Moreover, we also presented an extension to generate textual explanations in an offline setting, obtaining leading results on \bddx{}.

From a higher perspective, we advocate for more transparent and interpretable driving systems. %, such that their inner workings are understandable by human users. 
%In the future, we want to adapt our framework to more complex trajectory prediction systems that rely on multiple input sensors.
In the future, we want to investigate the possibility of generating natural language explanations in an online fashion, thus moving towards explanation-centric human-machine dialog.
\bibliographystyle{abbrvnat}
\bibliography{biblio}

\newpage

\section{Supplementary material}

\subsection{Details and additional results on the \hdd{} dataset}

\subsubsection{Implementation details}
Following previous work with \hdd{} \cite{hdd_cvpr2018,xu2019temporal,li2020learning}, we sample the videos at 3 Hz. Each frame is resized to $90 \times 160$ and concatenated with its $T=20$ previous frames to form a $(T+1) \times 3 \times 90 \times 160$ tensor. This sequence of images is passed as the input of a R(2+1)D convolutional neural network \cite{tran2018closer}, pretrained on Kinetics \cite{CarreiraZ17}. 
The driver is trained to predict the trajectory of the vehicle in the next 2 seconds. The scale of trajectories is in meters.
Besides, we found that we can increase driving performance by also learning to predict the positions two seconds \textit{in the past}, in addition to the two seconds \textit{in the future}. Thus, at every time step, we predict in total $K=13$ positions, corresponding to the local trajectory with 2 seconds horizon, at 3 Hz.
As no official validation split is provided for HDD, we validate our hyper-parameters on 10\% of the official training set that was kept aside training. The validation split that we used will be released along with our code.
The explanation module \acro{} merges the predicted trajectory with the internal driver representation, extracted at layer $L=3$, using a BLOCK fusion with a projection dimension of 256 and a core tensor composed of 5 blocks.
Once validated, we train our models on the full training set for 70K iterations and report the results on the official test split, similarly to \cite{li2020learning}.
As $\tens{L}_\textit{drive}$ and $\tens{L}_\textit{explain}$ are differentiable with respect of the parameters of the system, they are jointly optimized with ADAM \cite{adam} with a learning rate of $1.0\times 10^{-4}$ and a batch size of $12$. The hidden dimension of the GRU to decode the trajectory is 256. Weights are initialized with a Xavier initialization. Random seeds for initialization and optimization are fixed at the beginning of each training for reproducibility, and will be given in our code repository. All our models are developed using PyTorch 1.3, and trained on a single Nvidia GeForce RTX 2080 Ti with 11Go of RAM.
 
\subsubsection{Visualizations}

\begin{figure}[t]
    \begin{subfigure}{0.9\linewidth}
    \includegraphics[width=\textwidth]{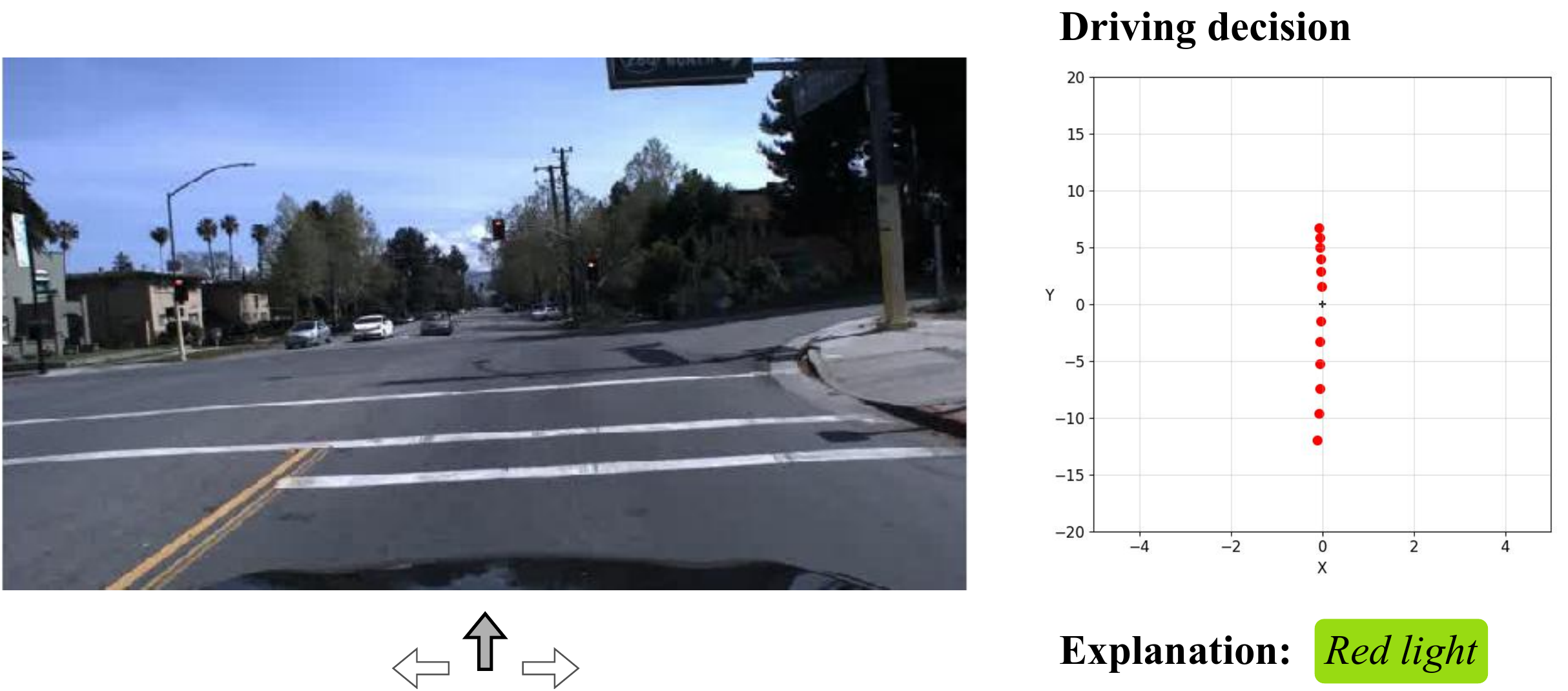}
    \caption{Example of a car stopping for a red light.}
    \label{fig:red_light}
    \end{subfigure}
    \begin{subfigure}{0.9\linewidth}
    \includegraphics[width=\linewidth]{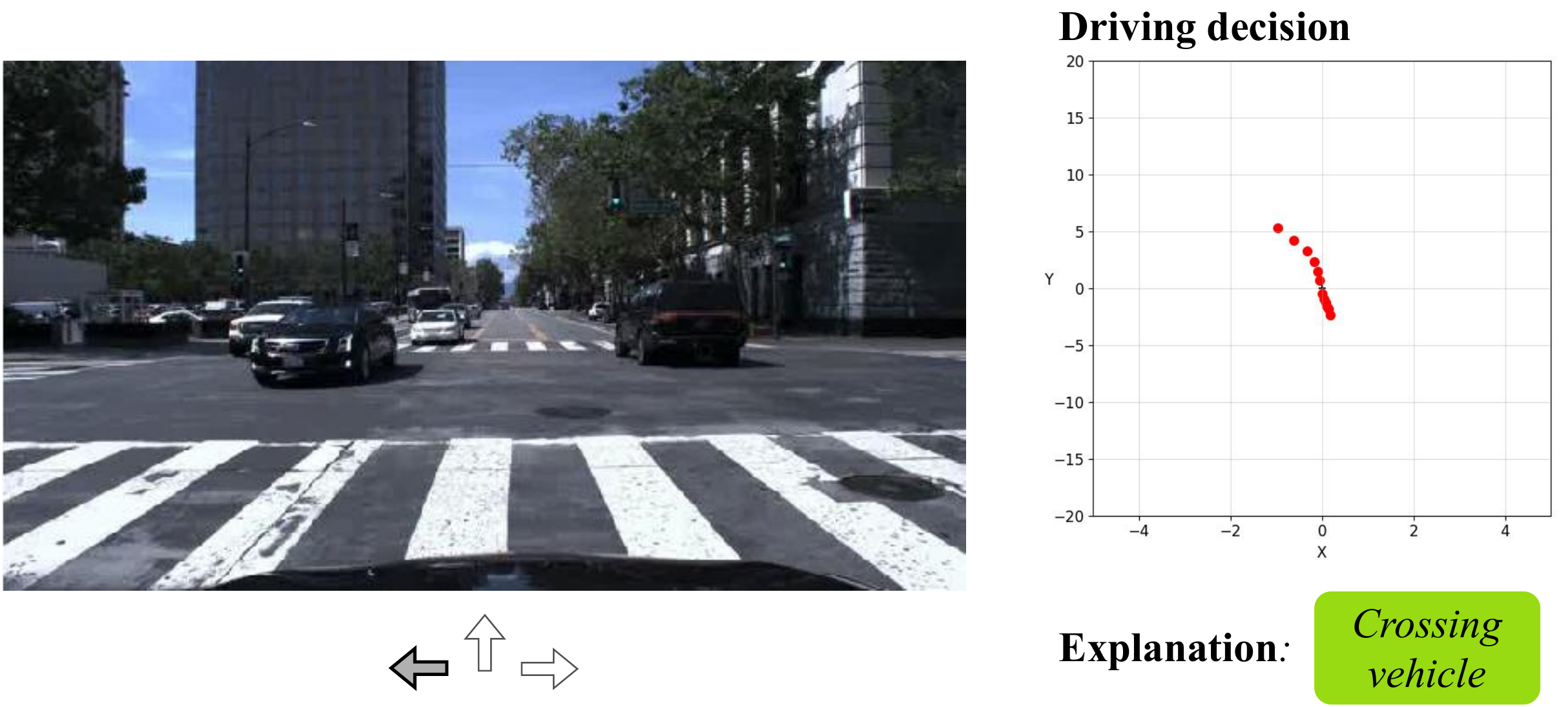}
    \caption{Example showing the importance of the high-level driver intent (blinker) to predict the future trajectory.}
    \label{fig:crossing_vehicle}
    \end{subfigure}
    \begin{subfigure}{0.9\linewidth}
    \includegraphics[width=\linewidth]{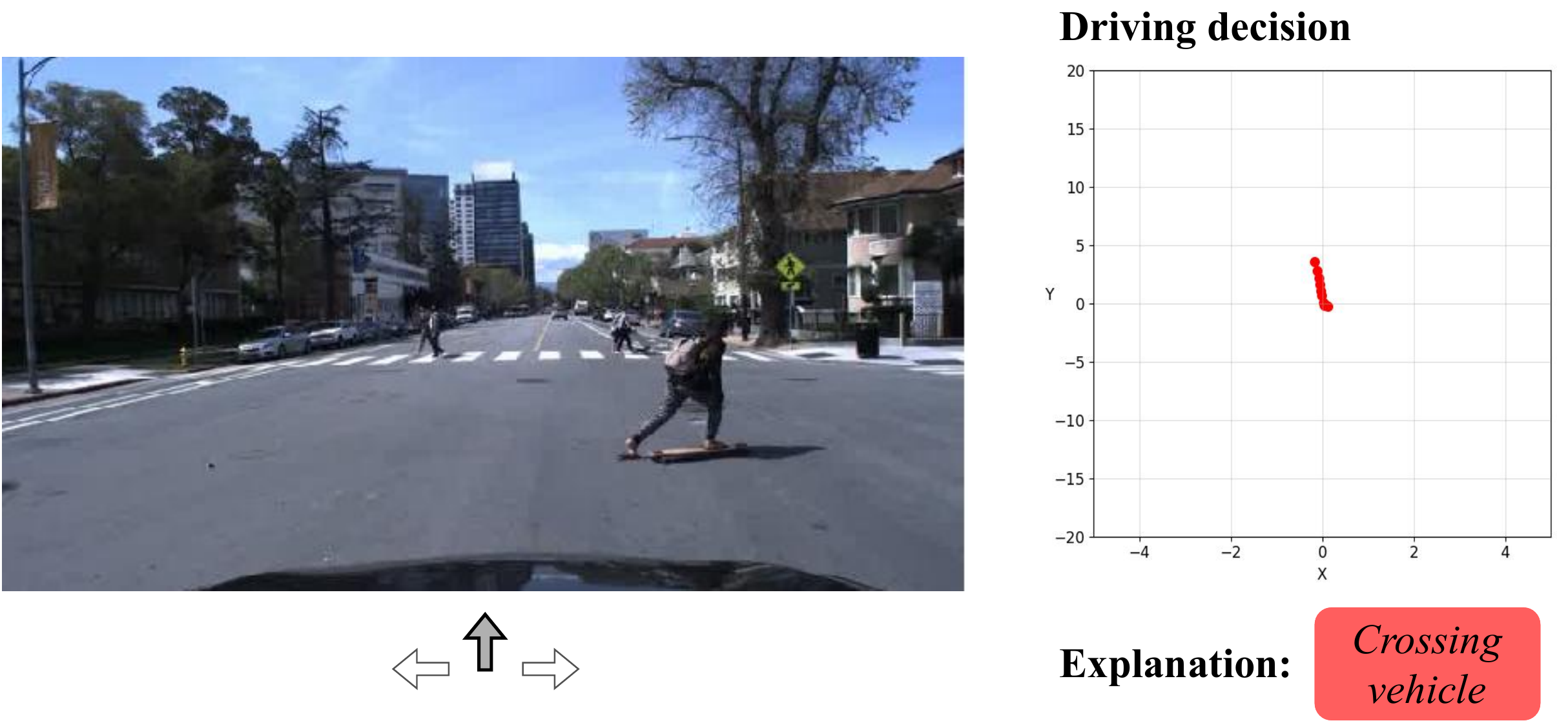}
    \caption{Example of a failure case where the person on the skateboard is identified as a crossing vehicle.}
    \label{fig:failure}
    \end{subfigure}
    \caption{Qualitative visualization of \acro{} predictions on the HDD dataset. Arrows under each camera image represent the blinker's state. For clarity's sake, only the image at time $t$ is shown here.}
    \label{fig:hdd_visu}
\end{figure}

\autoref{fig:hdd_visu} presents qualitative visualization of the output of \acro{} on test driving sessions of HDD.
In addition, we provide additional video visualizations of BEEF performances on HDD in the companion zipped folder.

%\subsection{Comparison to other fusion methods}
%In addition to the comparison that we report in the main paper, we provide in \autoref{tab:fusion_choice} the results of \acro{} with other fusion techniques, namely MLB \cite{DBLP:conf/iclr/KimOLKHZ17}, MFB \cite{mfb} and MFB \cite{mfh}.

\subsection{Details and additional results on the \bddx{} dataset}

\subsubsection{Implementation details}
We follow the same data preprocessing and experimental protocol as the previous work on BDD-X \cite{bddx}. In particular, we filter out explanations longer than 20 words from the training set and images are rescaled to $90\times160\times3$. The driving backbone is pretrained at 10 Hz and the explanation module attends over features of 20 frames equally spaced throughout the sequence.
Hyper-parameters were found on the provided validation set: batch size is 32, the learning rate of the ADAM optimizer \cite{adam} is $3\times10^{-4}$, the hidden dimension of the LSTM is 64, and layer 2 is used in the BLOCK fusion along with the decision features.
Our pretrained driver has 0.43 and 9.11 Mean Absolute Error (MAE) for acceleration and course values respectively. Training a model takes about 10 hours on an NVIDIA 2080Ti.

\subsubsection{Additional samples}

\autoref{tab:bdd_sample_extended} shows generated explanations (with a greedy decoding) on seven randomly chosen videos of the test set.
After a manual verification, we observe that 29 out of the 34 explanations correspond to the driving sequence and correctly explain the driving behavior.
Inspecting failures shows that they occur either by hallucinating street furniture (a stop light or a green light) or by interpreting the traffic as slowing down while there is no traffic (in one case).

In addition, we provide additional video visualizations of BEEF performances on BDD-X in the companion zipped folder.

%\subsection{Limits of natural language explanations}
%
%We observe two drawbacks inherent to the generation of natural language:
%
%\begin{itemize}
%    \item \textbf{Limits of the metrics.} Evaluating generated natural language samples is known to be a challenging problem \cite{bleu_problem_06}. The \bleu and \meteor metrics cannot solely measure the quality of the generated explanations \cite{bleu_problem_2017}. Moreover, \autoref{tab:bdd_sample_extended} shows that the scores do not always correlate with the exactitude of the explanations. 
%    Future work could leverage recent advances in evaluation of natural language generation \cite{DBLP:conf/naacl/HashimotoZL19,towards}.
%    
%    \item \textbf{Tradeoff between diversity and scores.} In \autoref{tab:bdd_sample_extended}, samples are generated with a greedy decoding technique, \ie the most likely word is sampled at each decoding step. More diversity can by obtained by sampling words according to the multinomial probability distribution (\ie using a higher temperature), but this comes at the expense of degraded \bleu and \meteor scores.
    
%\end{itemize}

\begin{table}[b]
    \centering
    \caption{\textbf{Explanations generated on BDD-X driving videos}. These videos come from the test set and \textit{are not ``cherry-picked''}. Explanations are generated by the \acro{} model and colors indicate whether the explanation matches the content of the video (\colorbox{green!20}{green}) or not (\colorbox{red!20}{red}), according to manual inspection. %: \textcolor{green}{green} means that the explanation is correct and \textcolor{red}{red} that the explanation does not match the content of the driving video.
    \bleu and \meteor metrics are abbreviated to \textbf{B4} and \textbf{M} respectively.}
    \resizebox{\textwidth}{!}{%
    %\begin{tabular}{@{}l l c c c@{}}
    \begin{tabular}{@{} l l l c c@{}}
        \toprule
        %\fontsize{11}{11}{\selectfont Human (gold-label)} & \fontsize{11}{11}{\selectfont Generated by \drivevoff{}} (greedy decoding) & Video & start & end \\
        %\fontsize{11}{11}{\selectfont Human (gold-label)} & \fontsize{11}{11}{\selectfont Generated by \drivevoff{}} (greedy decoding) \\
        & \fontsize{11}{11}{\selectfont Human (gold-label)} & \fontsize{11}{11}{\selectfont Generated by \acro{}} (greedy decoding) & \textbf{B4} & \textbf{M}\\
        \hline

1	&	\cellcolor{green!20}	since there are many obstacles to be aware of	&	\cellcolor{green!20}	because traffic is moving at a steady speed	&	0	&	0				\\
2	&	\cellcolor{green!20}	since the light ahead became red	&	\cellcolor{red!20}	because the light turned green	&	0	&	18.4				\\
3	&	\cellcolor{green!20}	since the light is red and there are people crossing the road	&	\cellcolor{green!20}	because the light is red	&	16.5	&	1.4				\\
4	&	\cellcolor{green!20}	because the lights are green and there is little traffic	&	\cellcolor{green!20}	because the road is clear	&	0	&	17.2				\\
5	&	\cellcolor{green!20}	because the traffic ahead starts moving faster	&	\cellcolor{green!20}	because traffic is moving forward	&	0	&	19.9				\\
6	&	\cellcolor{green!20}	because there are no nearby cars in its lane and the light is green	&	\cellcolor{green!20}	because the road is clear	&	0	&	12.5				\\
7	&	\cellcolor{green!20}	as it prepares to make a left hand turn	&	\cellcolor{red!20}	because there is a stop sign	&	0	&	2.2				\\
8	&	\cellcolor{green!20}	to enter another road while the light is green	&	\cellcolor{green!20}	because the road is clear	&	0	&	15.2				\\
9	&	\cellcolor{green!20}	because there are no nearby cars in its lane impeding it	&	\cellcolor{green!20}	because the road is clear	&	0	&	5.9				\\
10	&	\cellcolor{green!20}	for the red light and a pedestrian in the crosswalk	&	\cellcolor{green!20}	because the light ahead is red	&	0	&	16.1				\\
11	&	\cellcolor{green!20}	for the red light	&	\cellcolor{green!20}	because the light is red	&	0	&	33.1				\\
12	&	\cellcolor{green!20}	because there are cars and people nearby	&	\cellcolor{green!20}	because traffic is moving at a steady speed	&	0	&	8.7				\\
13	&	\cellcolor{green!20}	because the car can turn right	&	\cellcolor{green!20}	because the road is clear	&	0	&	15.2				\\
14	&	\cellcolor{green!20}	to avoid the car in front on the left	&	\cellcolor{green!20}	because the road is clear	&	0	&	2.5				\\
15	&	\cellcolor{green!20}	because there are pedestrians blocking the path to the right	&	\cellcolor{red!20}	because there is a stop sign	&	0	&	23.2				\\
16	&	\cellcolor{green!20}	since it is waiting for the pedestrians to cross	&	\cellcolor{red!20}	because there is a stop sign	&	0	&	3.5				\\
17	&	\cellcolor{green!20}	because the pedestrians had moved out of the way	&	\cellcolor{green!20}	because the road is clear	&	0	&	16.9				\\
18	&	\cellcolor{green!20}	since the car had just turned right	&	\cellcolor{green!20}	because the road is clear	&	0	&	4				\\
19	&	\cellcolor{green!20}	because traffic is now moving in both lanes	&	\cellcolor{green!20}	because traffic is moving forward	&	0	&	25.2				\\
20	&	\cellcolor{red!20}	because traffic is moving forward	&	\cellcolor{green!20}	because the car in front is slowing down	&	0	&	11.6				\\
21	&	\cellcolor{green!20}	because traffic is moving faster in that lane	&	\cellcolor{green!20}	because the lane is clear	&	0	&	16.5				\\
22	&	\cellcolor{green!20}	because now that lane is moving faster	&	\cellcolor{green!20}	because the car in front is slowing down	&	0	&	9.3				\\
23	&	\cellcolor{green!20}	because traffic is moving slowly	&	\cellcolor{green!20}	because traffic is moving at a steady speed	&	34.6	&	39.7				\\
24	&	\cellcolor{green!20}	because the cars ahead have stopped	&	\cellcolor{green!20}	because the light is red	&	0	&	14.1				\\
25	&	\cellcolor{green!20}	because the van in front was'nt moving	&	\cellcolor{green!20}	because traffic is stopped	&	0	&	7.1				\\
26	&	\cellcolor{green!20}	since the van started moving	&	\cellcolor{green!20}	because traffic is moving slowly	&	0	&	14.1				\\
27	&	\cellcolor{green!20}	because the van stopped moving	&	\cellcolor{green!20}	because traffic is stopped	&	0	&	27.2				\\
28	&	\cellcolor{green!20}	because the van in front is moving at different speeds	&	\cellcolor{green!20}	because traffic is moving slowly	&	0	&	15.1				\\
29	&	\cellcolor{green!20}	as traffic in front of it moves slowly and merges together	&	\cellcolor{green!20}	because the car in front of it is moving forward	&	23.8	&	14.5				\\
30	&	\cellcolor{green!20}	when traffic in front of it stops	&	\cellcolor{green!20}	because the car in front of it is stopped	&	29.9	&	24.1				\\
31	&	\cellcolor{green!20}	as traffic in front of it moves forward slightly	&	\cellcolor{green!20}	because the car in front of it has stopped	&	29.9	&	17.5				\\
32	&	\cellcolor{green!20}	because there are no nearby cars in its lane	&	\cellcolor{red!20}	because the car in front of it has slowed down	&	0	&	15.7				\\
33	&	\cellcolor{green!20}	while there is a gap in traffic	&	\cellcolor{green!20}	because the car in front is moving forward	&	0	&	5.9				\\
34	&	\cellcolor{green!20}	because there are no nearby cars in front of it	&	\cellcolor{green!20}	because the car in front is moving forward at a normal speed	&	0	&	22.8				\\

        \bottomrule
    \end{tabular}
    }
    \label{tab:bdd_sample_extended}
\end{table}

\end{document}